\documentclass[letterpaper, 10 pt, conference]{ieeeconf}  % Comment this line out if you need a4paper
\IEEEoverridecommandlockouts                              % This command is only needed if
\usepackage{graphicx} % for pdf, bitmapped graphics files
\usepackage{times} % assumes new font selection scheme installed
\usepackage{cite}
\usepackage{amsmath} % assumes amsmath package installed
\usepackage{amssymb}  % assumes amsmath package installed
\usepackage{amsfonts}       % blackboard math symbols
\usepackage{dsfont}
\usepackage{braket}
\usepackage{booktabs}
\usepackage{algorithm,algorithmic}
\usepackage{color}
\usepackage{colortbl}
\usepackage{url}
\usepackage{mathtools}
\usepackage{multirow}
\usepackage{footmisc}

\usepackage[caption=false, font=footnotesize]{subfig}

\usepackage{lipsum}
\usepackage{widetext}

\newcommand{\obs}{\boldsymbol{x}}
\newcommand{\action}{\boldsymbol{a}}

\newcommand{\latent}{{\boldsymbol z}}

\newcommand{\latenth}{{\boldsymbol h}}

\newcommand{\best}[2]{\underline{\textbf{#1}}$\pm$\scriptsize{#2}}
\newcommand{\second}[2]{\underline{#1}$\pm$\scriptsize{#2}}
\newcommand{\third}[2]{#1$\pm$\scriptsize{#2}}

\title{\LARGE \textbf{
DreamingV2: Reinforcement Learning \\ with Discrete World Models without Reconstruction
}}

\author{Masashi Okada$^{\dag,\star}$ and Tadahiro Taniguchi$^{\dag,*}$% <-this % stops a space
\thanks{$^{\dag}$ Masashi Okada and Tadahiro Taniguchi are with Digital \& AI Technology Center, Technology Division, Panasonic Corporation, Japan.
}%
\thanks{$^{*}$ Tadahiro Taniguchi is also with Ritsumeikan University, College of Information Science and Engineering, Japan.
}%
\thanks{$^{\star}$ \texttt{okada.masashi001@jp.panasonic.com}
}
}

\newcommand{\minibatch}{\mathcal{D}}

\newcommand{\nceloss}{\mathcal{J}^{\mathrm{NCE}}}
\newcommand{\klloss}{\mathcal{J}^{\mathrm{KL}}}
\newcommand{\llloss}{\mathcal{J}^{\mathrm{likelihood}}}

\newcommand{\proposedmethod}{DreamingV2}

\begin{document}

\maketitle
\thispagestyle{empty}
\pagestyle{empty}

%%%%%%%%%%%%%%%%%%%%%%%%%%%%%%%%%%%%%%%%%%%%%%%%%%%%%%%%%%%%%%%%%%%%%%%%%%%%%%%%
\begin{abstract}
The present paper proposes a novel reinforcement learning method with world models, DreamingV2, a collaborative extension of DreamerV2 and Dreaming.
DreamerV2 is a cutting-edge model-based reinforcement learning from pixels that uses discrete world models to represent latent states with \textit{categorical variables}.
% DreamerV2 is an attractive solution for robotic learning because its discrete representation should be suitable to describe contact-rich and discontinuous environments.
% However, the world models are based on variational autoencoding, and reconstructing complex visual observations has been a bottleneck of training.
Dreaming is also a form of reinforcement learning from pixels that attempts to avoid the autoencoding process in general world model training by involving a \textit{reconstruction-free} contrastive learning objective.
% However, likewise earlier world models literature, Dreaming employs continuous latent states and does not take advantage of discrete representation.
The proposed DreamingV2 is a novel approach of adopting both the discrete representation of DreamingV2 and  the reconstruction-free objective of Dreaming.
Compared to DreamerV2 and other recent model-based methods without reconstruction, DreamingV2 achieves the best scores on five simulated challenging 3D robot arm tasks.
We believe that DreamingV2 will be a reliable solution for robot learning since its discrete representation is suitable to describe discontinuous environments, and the reconstruction-free fashion well manages complex vision observations.
\end{abstract}

%%%%%%%%%%%%%%%%%%%%%%%%%%%%%%%%%%%%%%%%%%%%%%%%%%%%%%%%%%%%%%%%%%%%%%%%%%%%%%%%
\section{Introduction} \label{sec:intro}
World models~\cite{ha2018world} are a potential approach to achieving the visual servoing of robots in the industry.
The world models, which are equipped with compact latent representation models and latent forward dynamics,
efficiently predict future trajectories and rewards, allowing us to acquire model predictive controllers~\cite{watter2015embed,hafner2018learning,okada2020planet} and policies learned by model-based reinforcement learning~\cite{hafner2019dreamer,hafner2020mastering,okada2021dreaming}.
In addition, world modes have various valuable properties for industrial applications, such as transferability to new tasks~\cite{byravan2020imagined}, unsupervised exploration~\cite {sekar2020planning}, generalization from offline datasets \cite{yu2020mopo}, and explainability~\cite{sakai2022explainable}.
World model based agents have demonstrated state-of-the-art results on a wide range of simulated tasks from pixels, such as DeepMind Control Tasks~\cite{deepmindcontrolsuite2018} and the Atari benchmark~\cite{bellemare2013arcade}, but there is still a scarcity of research on real-world applications.
\begin{figure}
  \centering
  \includegraphics[width=0.48\textwidth]{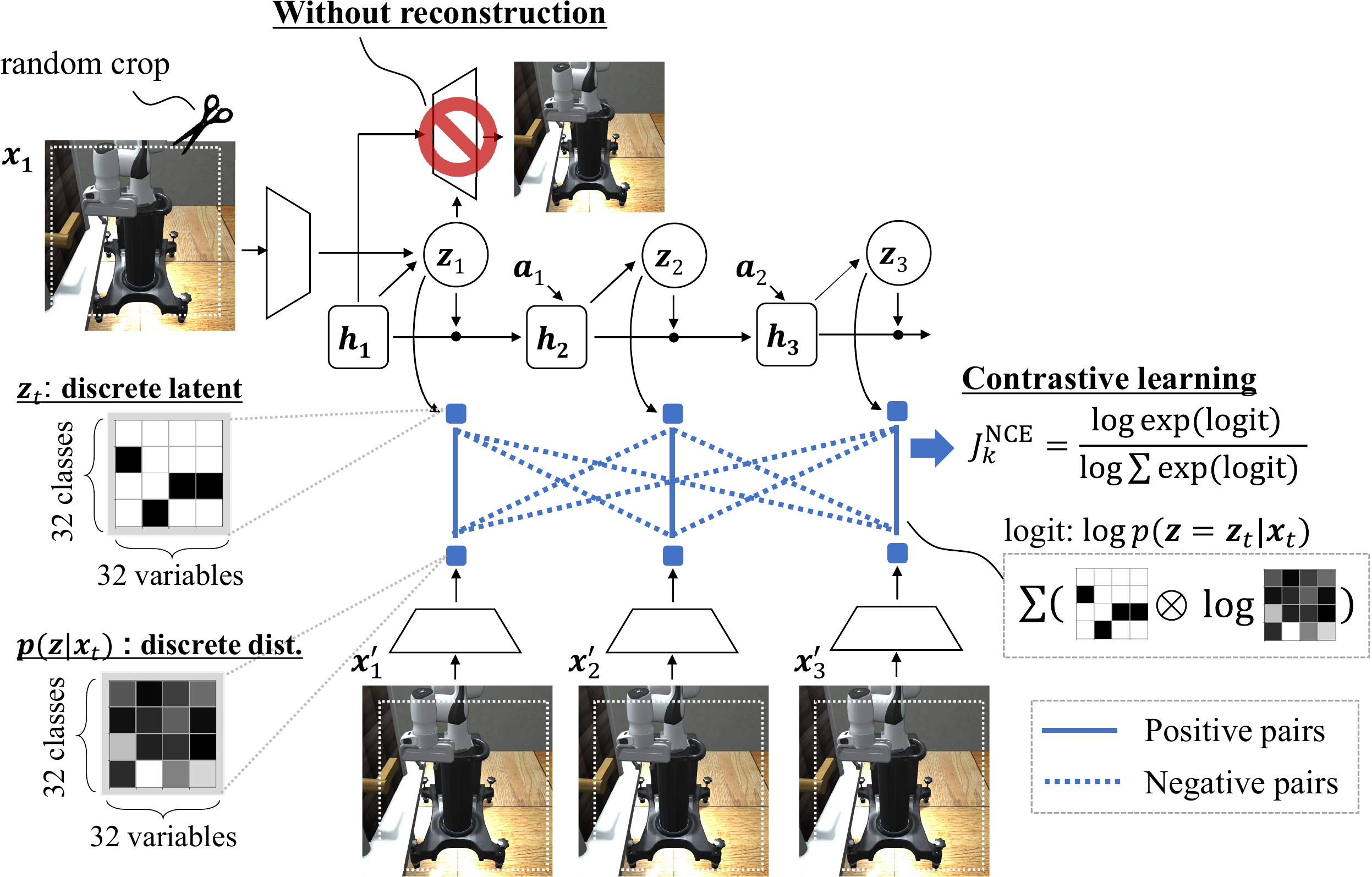}
  \caption{
    DreamingV2's contrastive learning to train discrete world models without reconstruction.
    The trained discrete world models can successfully represent simulated 3D robot arm environments.
    Given an observation $\obs_{t}$, an inference model $q(\latent_{t}|\latenth_{t}, \obs_{t})$ and generative model $p(\latent_{t}|\latenth_{t})$ predict trajectories of discrete latent states $\latent_{t}$.
    While, other inference model predicts probability mass functions $p(\latent|\obs_{t})$ from time series of observations $\obs_{t:t+K}$.
    The predicted latents and probability mass functions are paired to compute logits, and then contrastive learning is conducted.
    As in Dreaming\cite{okada2021dreaming}, random crop image augmentation is exploited.
  } \label{fig:fig1}
\end{figure}

DreamerV2~\cite{hafner2020mastering} is a leading type of world model based reinforcement learning that achieved human-level performance on the Atari benchmark.
Unlike previous world models~\cite{watter2015embed,hafner2018learning,okada2021dreaming}, including Dreamer~\cite{hafner2019dreamer} (the earlier version of DreamerV2), this method uses discrete world models in which discrete random variables represent latent states.
A motivation to introduce discrete representation is that categorical distributions can naturally capture multimodal uncertainty of stochastic state transitions.
In contrast, earlier world models that use Gaussian distribution cannot manage multimodal uncertainty well~\cite{okada2019variational}.
The discrete state representation would be useful for real-world robot tasks for the following reasons.
First, practical robot tasks demand contact-rich manipulation and making the dynamics highly discontinuous.
Second, several robot tasks consist of subtasks.
Let us consider a door opening task by robot arms, which includes the subtasks of moving the end-effector closer to a door handle,
grabbing the handle, and then pulling it to open the door.
To represent such unsmooth state space and discrete phase changes, the use of discrete variables is quite reasonable.

However, DreamerV2 learns the world models using a general time-series variational auto-encoding objective~\cite{hafner2018learning}, and reconstruction process of complex visual observations will produce various issues in practical circumstances.
For example, as pointed out in \cite{okada2021dreaming}, the autoencoder often fails to recognize small task-relevant objects, which is referred to as \textit{object vanishing}.
In addition, the autoencoder-based world models are sensitive to visual distraction such as unseen task-irrelevant objects and noises like shadows~\cite{ma2020contrastive,nguyen2021temporal,deng2021dreamerpro}.

Dreaming~\cite{okada2021dreaming}, CVRL (Contrastive Variational Reinforcement Learning)~\cite{ma2020contrastive}, TPC (Temporal Predictive Coding)~\cite{nguyen2021temporal}, and DreamerPro~\cite{deng2021dreamerpro} are another type of world model methods that involves reconstruction-free objectives to solve the above issues produced by the autoencoding.
% Dreaming~\cite{okada2021dreaming}, CVRL (Contrastive Variational Reinforcement Learning)~\cite{ma2020contrastive}, TPC (Temporal Predictive Coding)~\cite{nguyen2021temporal}, and DreamerPro~\cite{deng2021dreamerpro}, have proposed reconstruction-free objective to train world models.
For instance, Dreaming is a reconstruction-free modification of Dreamer~\cite{hafner2019dreamer} that uses self-supervised contrastive learning~\cite{oord2018representation,chen2020simple}.
However, these methods require a continuous state variable and do not benefit from the discrete representation described above.

% A solution to the autoencoding related problems is to remove the decoder
% by incorporating self-supervised schemes.
% Dreaming (\textit{\underline{Dream}er with \underline{In}foMax and without \underline{g}enerative decoder})~\cite{okada2021dreaming}, TPC (temporal predictive coding)~\cite{nguyen2021temporal}, DreamerPro~\cite{deng2021dreamerpro},
% have proposed decoder-free world model learning schemes.
% However, these method mainly dedicated to continuous latent variable so that
% they cannot handle the discrete latent states.
% %
% Motivated by this, we propose a DreamingV2 based on Dreaming.

Motivated by the above, we propose a DreamingV2 by exploiting and merging the main principles of DreamerV2 and Dreaming.
The top-level concept is summarized in Fig.~\ref{fig:fig1}.
Our primary contributions are summarized as follows:
\begin{itemize}
  \item We devise a contrastive learning approach for training discrete world models without reconstruction.
  To the best of our knowledge, it is the first application of contrastive learning for categorical discrete representations.
  \item We demonstrate that DreamingV2, reinforcement learning using the discrete world models, can achieve  state-of-the-art results on five simulated robot-arm tasks.
\end{itemize}
The remainder of this paper is organized as follows.
In Sec.~\ref{sec:preliminary}, we present an overview of Dreamer, DreamerV2, and Dreaming.
In Sec.~\ref{sec:proposed_method}, our proposed method DreamingV2 is specified.
In Sec.~\ref{sec:experiments}, the effectiveness of \proposedmethod{} is demonstrated via simulated evaluations.
In Sec.~\ref{sec:related_work}, related work are summarized.
Finally, Sec.~\ref{sec:conclusion} concludes this paper.

\begin{figure*}
  % \begin{minipage}{0.55\textwidth}
    \centering
    \includegraphics[width=0.55\textwidth]{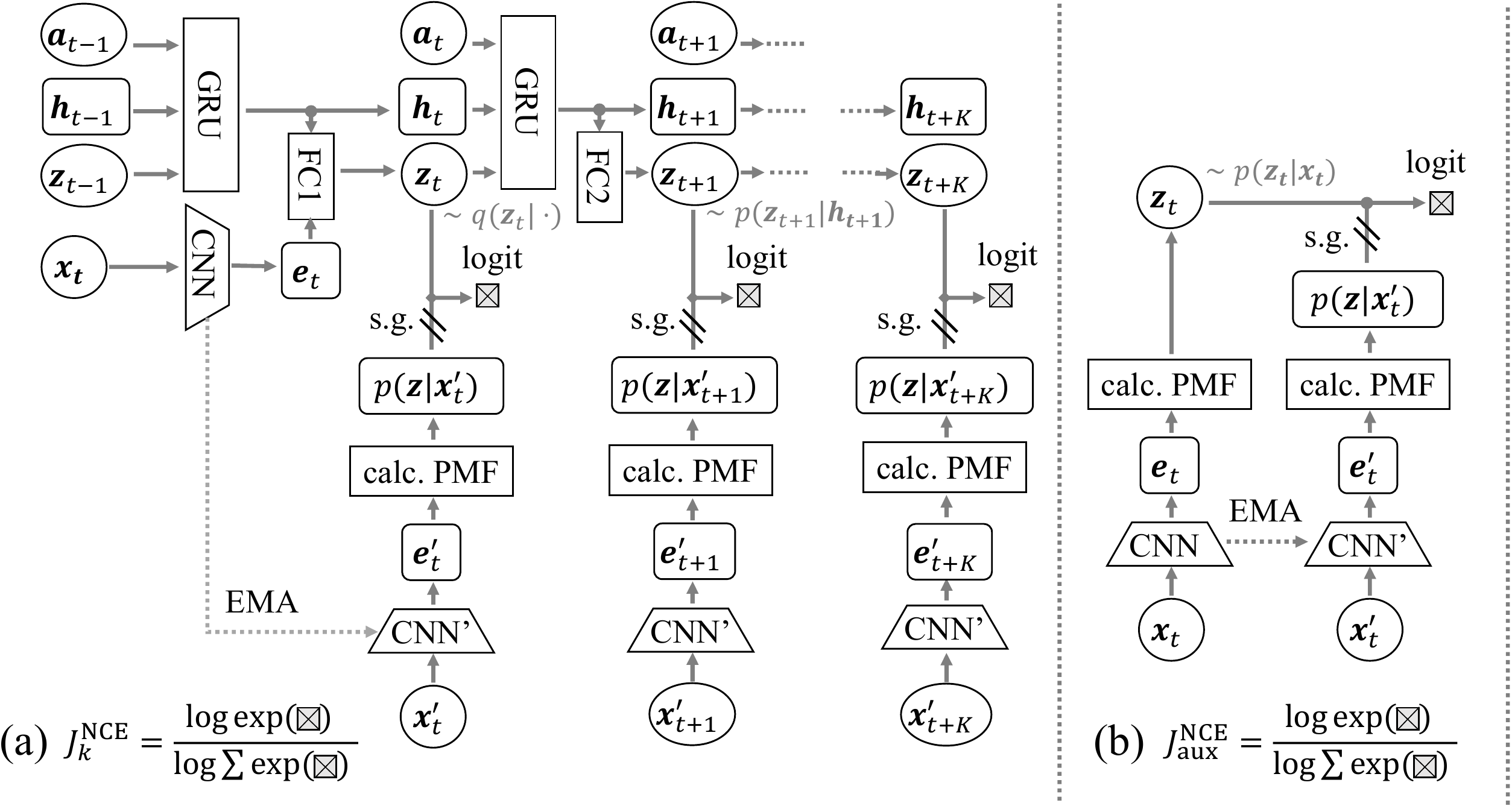}
    \hspace*{0.5cm}
    \includegraphics[width=0.3\textwidth]{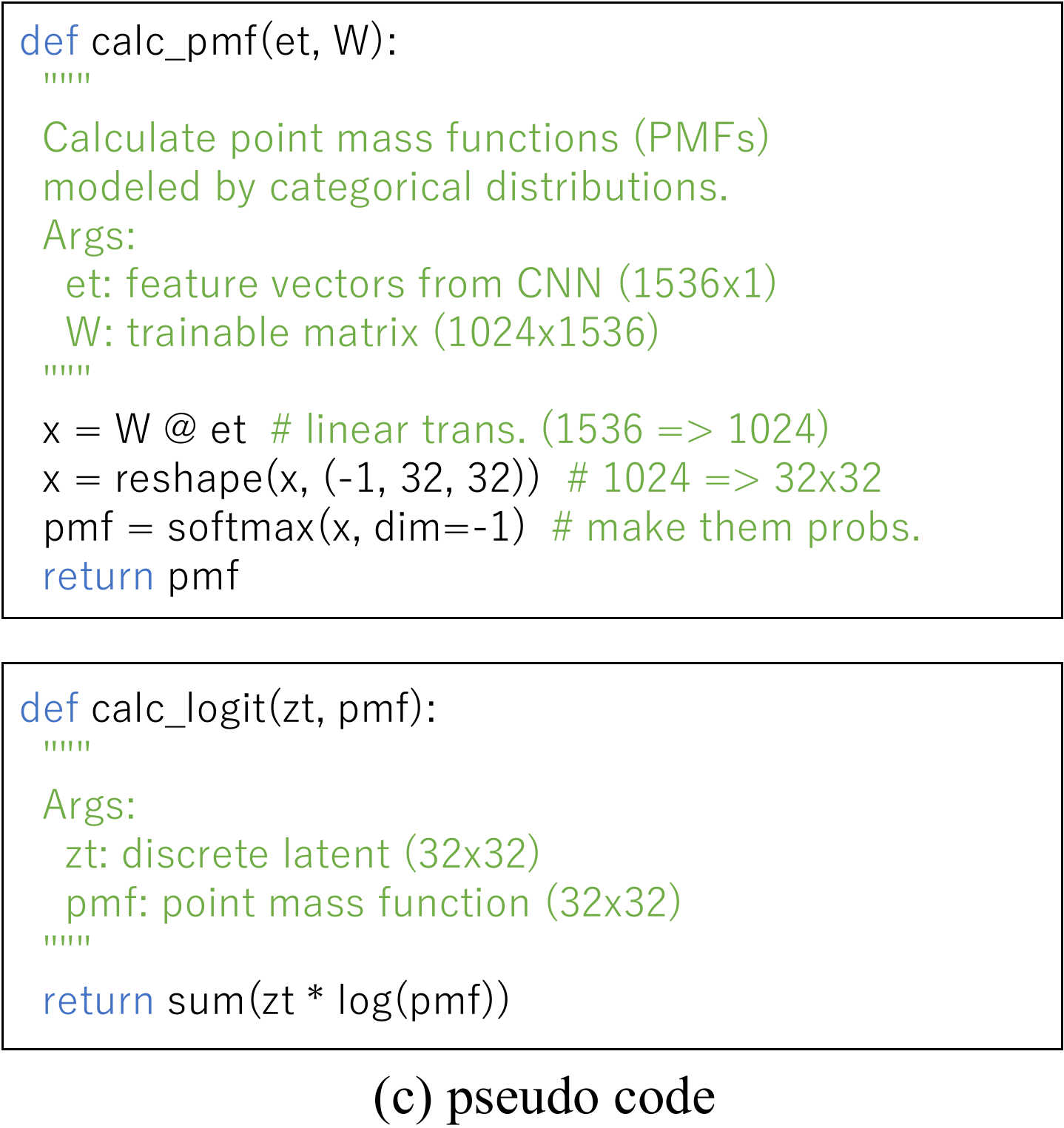}
    \caption{
      The RSSM-based architecture and PyTorch-like pseudocode to compute $\nceloss_{k}$ and $\nceloss_{\mathrm{aux}}$.
      \texttt{CNN}, \texttt{GRU}, and \texttt{FC} represent a convolutional neural network, a GRU-cell, and a fully-connected layer, respectively.
      s.g. and EMA are short for stop-gradient and exponential moving average.
      (a) To compute $\nceloss_{k}$, $\latent_{t:t+K}$ are inferred and predicted using $q(\latent_{t})$ and $p(\latenth_{t+1}|\latent_{t})$ from an observation $\obs_{t}$, where $K$ is the prediction horizon.
      The predicted latent are compared with the point mass functions estimated from observations $\obs'_{t:t+K}$ and then $B\times(K+1)$ logits are computed, where $B$ is the number of episodes in the mini-batch.
      The notations of $\obs_{t}$ and $\obs'_{t}$ indicate the different random augmentation in Sec.~\ref{sec:random_aug} is applied.
      For readability, we only illustrate positive logits, however, negative logits are also computed by pairwise comparison of the latent from different time-steps and episodes, yielding $(B\times(K+1))^{2}$ logits.
      (b) To compute $\nceloss_{\mathrm{aux}}$, only the encoders and critic $p(\latent|\obs_{t})$ are taken into account. Also in this figure, we only illustrate positive logits.
      }
    \label{fig:arch_and_code}
  % \end{minipage}
\end{figure*}%

\section{Preliminary} \label{sec:preliminary}
This section briefly reviews the Recurrent State Space Model (RSSM),
and RSSM based reinforcement learning methods: Dreamer, DreamerV2, and Dreaming.

\subsection{Recurrent State Space Model}
RSSM is a principal model for representing world models, and is used as a major component of various methods~\cite{hafner2018learning,hafner2019dreamer,han2019variational,okada2020planet,sekar2020planning,okada2021dreaming,deng2021dreamerpro,hafner2020mastering}.
RSSM defines the latent variable as a tuple of $(\latent_{t}, \latenth_{t})$
where $\latent_{t}$, $\latenth_{t}$ are the stochastic and deterministic variables, respectively.
RSSM's generative and inference models are defined as the following:
\begin{align}
  \mathrm{Generative\ models}&:
  \begin{cases}
    \latenth_{t} = f^{\mathrm{GRU}}(\latenth_{t-1}, \latent_{t-1}, \action_{t-1}) \\
    \latent_{t} \sim p(\latent_{t} | \latenth_{t}) \\ \label{eqn:rssm}
    \obs_{t} \sim p(\obs_{t} | \latenth_{t}, \latent_{t})
  \end{cases}, \\\nonumber
  \mathrm{Inference\ model}&: \latent_{t} \sim q(\latent_{t} | \latenth_{t}, \obs_{t}), 
  % s_{t} &\sim 
\end{align}
where $\obs$ and $\action$ denote pixel observations and actions, respectively.
The deterministic $\latenth_{t}$ is considered to be the hidden state of the gated recurrent unit (GRU) $f^{\mathrm{GRU}}(\cdot)$~\cite{cho2014learning}. % so that temporal information are embedded into $\latenth_{t}$.
Generally, these models are trained by optimizing the variational evidence lower bound objective (ELBO):
\begin{align}
  &\mathcal{J}^{\mathrm{ELBO}} = \sum_{t}\left(
  \underbrace{
  \mathbb{E}_{q(\latent_{t}|\latenth_{t}, \obs_{t})}\left[
    \log p(\obs_{t}|\latenth_{t}, \latent_{t})
  \right]
  }_{\coloneqq \llloss}
  \right.
  \label{eqn:vae_elbo}\\
  & \left.
  \underbrace{
  - \mathbb{E}_{q(\latent_{t}|\cdot)}\left[
    {\operatorname{KL}}\left[
      q(\latent_{t+1} |\latenth_{t+1}, \obs_{t+1}) ||
      p(\latent_{t+1} |\latenth_{t+1})
  \right]
  \right]
  }_{\coloneqq \klloss}
  \right). \nonumber
\end{align}
To conduct model predictive control or policy optimization, the reward predictor $p(r_{t}|\latenth_{t}, \latent_{t})$ is also necessary, but it is simply realized by regarding the rewards as observations and learn the reward predictor along with the generative model $p(\obs_{t}|\latenth_{t},\latent_{t})$.
For readability, the following discussion omits the specifications of the reward function.
% $\llloss$ is an reconstruction objective to assure latent embeddings have enough information to reconstruct pixels.
% $\klloss$ trains the dynamics $p(\latent_{t}|\latenth_{t})$ toward the inference model $q(\latent_{t}|\cdot)$, and it regularizes the representations toward the prior.

\subsection{Dreamer and DreamerV2}
Dreamer and DreamerV2~\cite{hafner2019dreamer,hafner2020mastering} use RSSM to imagine latent trajectories and future rewards to train their policies.
The training procedure of Dreamer(V2) is simply summarized as follows:
{(i)} train RSSM with a given replay buffer by optimizing Eq.~(\ref{eqn:vae_elbo}),
{(ii)} train a policy within the trained RSSM-based world model by latent imagination, and
{(iii)} execute the trained policy in a real environment and store the observed results with the replay buffer.
The steps above are iteratively performed until some termination condition is satisfied.
The major difference between Dreamer and DreamerV2 is the specification of the stochastic latent $\latent_{t}$;
Gaussian random variables vs.~categorical random variables.
DreamerV2 adopted other minor modifications (e.g., KL-balancing and policy entropy regularization), but they are not detailed in this paper since the proposed method adopts the components as-is.

\subsection{Dreaming}
Dreaming~\cite{okada2021dreaming} introduces a reconsruction-free objective derived from the ELBO objective:
\begin{align}
  \mathcal{J}^{\mathrm{Dreaming}} \coloneqq \textstyle\sum^{K}_{k=0} \left(\nceloss_{k} + \klloss_{k}\right), \label{eqn:dreaming_objective}
\end{align}
where $K$ is the prediction horizon, and $\nceloss_{k}$ and $\klloss_{k}$ are referred to as Info-NCE (noise contrastive estimator) objective and overshooting objective, respectively.
Since $\nceloss_{k}$ mainly characterizes Dreaming, enabling to train RSSM without reconstruction, this section only describes this objective. $\nceloss_{k}$ is defined as:
\begin{align}
  & \nceloss_{k} \coloneqq \mathbb{E}_{\tilde{p}(\latent_{t}|\latent_{t-k},\action_{<t})q(\latent_{t-k}|\cdot)}\left[
      % \log p(\latent_{t}|\obs_{t}) - \log \sum_{\obs'} p(\latent_{t}|\obs')
      \log\frac{p(\latent_{t}|\obs_{t})}{\sum_{\obs'\in \mathcal{D}} p(\latent_{t}|\obs')}
  \right], \nonumber
\end{align}
where $\tilde{p}(\latent_{t}|\cdot)$ and $p(\latent_{t}|\obs_{t})$ are respectively referred to as \textit{auxiliary dynamics} and \textit{critic} in this paper, and $\mathcal{D}$ indicates a mini-batch.
The auxiliary dynamics is responsible for predicting $\latent_{t}$ from $\latent_{t-k}$, and critic is to evaluate $\latent_{t}$ predicted by the auxiliary dynamics.
The definitions of the auxiliary dynamics and critic are introduced later in Sec.~\ref{sec:proposed_method} to highlight the difference from DreamingV2.
Let $|\mathcal{D}|$ be the size of $\minibatch$, $\nceloss_{k}$ is considered as a $|\mathcal{D}|$-class categorical cross entropy objective to discriminate the positive pair $(\latent_{t}, \obs_{t})$ among the other negative pairs $(\latent_{t}, \obs'(\neq\obs_{t}))$.
Representation learning with this type of objective is known as contrastive learning~\cite{oord2018representation,chen2020simple}.
This objective can be easily computed by softmax cross entropy functions, generally predefined in deep learning frameworks%
\footnote{
  i.e., \texttt{softmax\_cross\_entropy\_with\_logits()} in TensorFlow, and \texttt{cross\_entropy()} in PyTorch.
},
by inputting $\log p(\latent|\obs)$ as \textit{logit}.

\begin{figure*}
  \centering
  \includegraphics[width=0.95\textwidth]{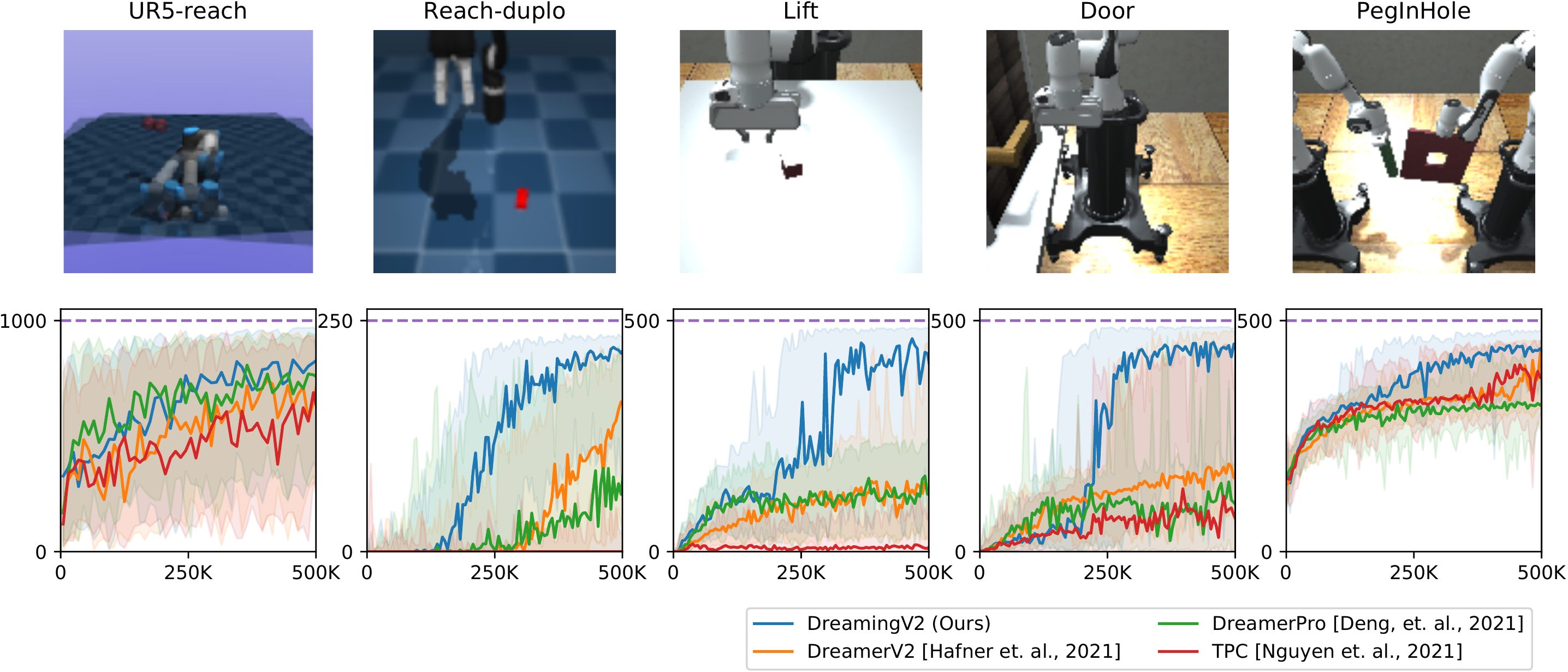}
  \caption{
    Simulated robot-arm tasks (top) and training curves (bottom).
    The UR5-reach is originally introduced in \cite{okada2021dreaming},
    the Reach-duplo task is from DeepMind Control Suite~\cite{deepmindcontrolsuite2018},
    and the remaining tasks are from RoboSuite~\cite{zhu2020robosuite} with iGibsonRenderer~\cite{li2021igibson} for photorealistic visualization.
    In the bottom figures, horizontal and vertical axes indicate time steps and episode reward, respectively.
    The lines represent the medians, and the shaded areas depict the percentiles 5\% to 95\% on 8 random seeds.
    The horizontal dashed lines indicate the upper limit return of the tasks.
  }
  \label{fig:main_result}
\end{figure*}

\section{DreamingV2} \label{sec:proposed_method}
This section introduces a new model-based reinforcement learning that uses discrete world models trained by contrastive learning, which we call DreamingV2.
Figure~\ref{fig:arch_and_code} illustrates the detailed architecture and PyTorch-like psuedo code of some essential operations.
DreamingV2's objective is almost similar to Dreaming as:
\begin{align}
  J^{\mathrm{DreamingV2}} \coloneqq J^{\mathrm{Dreaming}} + \nceloss_{\mathrm{aux}}.
\end{align}
A difference between DreamingV2 with Dreaming is an auxiliary objective $\nceloss_{\mathrm{aux}}$ (defined later in Sec.~\ref{sec:aux_obj}).
The other significant difference of DreamingV2 from Dreamer is that continuous latent variables and their distributions have been replaced with discrete variables and distributions.
% the auxiliary dynamics $\tilde{p}(\latent_{t}|\cdot)$ and critic $p(\latent_{t}|\obs_{t})$ are defined with discte distributions.
% In the fo
In the following sections, the detail of DreamingV2 is specified, with focus on the changes from Dreaming.

\subsection{Latent state representation}
DreamingV2 replaces continuous latent variables with discrete variables in a similar format to DreamerV2.
The stochastic latent state $\latent$ comprises 32 categorical variables, and each variable is a 32-dimensional one-hot vector.
With this specification, all distributions of $\latent$ are point mass functions (PMFs) modeled by categorical distributions.
The PMF is parameterized with a $32 \times 32$ probability matrix.
To backpropagate through the discrete latent sampled from the categorical distributions, straight-through gradients~\cite{bengio2013estimating} are involved as in DreamerV2.

\subsection{Definition of auxiliary dynamics}
Dreaming defined the auxiliary dynamics $\tilde{p}(\latent_{t}|\cdot)$ as a trainable linear dynamics.
An intuitive motivation to introduce such a simple model was that linear dynamics successfully constrain temporally consecutive latent states (namely negative pairs in contrastive learning) are not distributed too further.
% In the literature of Dreaming, this linear dynamics was regarded a key component to achieve the state-of-the-art results.
Instead, DreamingV2 defines the auxiliary dynamics using the RSSM generative models $p(\latent_{t}|\latenth_{t}=f^{\mathrm{GRU}}(\cdot))$ in Eq.~(\ref{eqn:rssm}).
In Fig.~\ref{fig:arch_and_code}(a), multi-step prediction is conducted using RSSM models.
Without the linear modeling, the discrete representation itself successfully regularizes \textit{repulsions} between the consecutive negative pairs.
This is experimentally supported in the ablation study in Sec.~\ref{sec:ablation_aux_dyn}.
% ,in which no significant difference is observed between RSSM and linear modeling.
% This result is contrary
% , since the latent state space itself successfully regularize the repulsions between negative pairs in contrastive learning.
% In addition, linear dynamics is less expressive to represent discrete state transitions.

\subsection{Definition of Critic} \label{sec:critic}
Dreaming defined the critic $p(\latent_{t}|\obs_{t})$ as:
\begin{align}
% $
  p(\latent_{t}|\obs_{t}) \propto \exp(\latent_{t}^{\top}W\boldsymbol{e}_{t}), \label{eqn:bilinear}
% $
\end{align}
where $W$ is a trainable matrix, $\boldsymbol{e}_{t} \coloneqq f^{\mathrm{CNN}}(\obs_{t})$, and $f^{\mathrm{CNN}}(\cdot)$ denotes feature extraction by a CNN unit.
Since $\log p(\latent_{t}|\obs_{t})$ is fed into the softmax cross entropy function, cosine similarity metrics $\latent_{t}^{\top}W\boldsymbol{e}_{t}$ are used as logits for contrastive learning.
It is interesting to note that Eq.~(\ref{eqn:bilinear}) is the von-Mises-Fisher distribution (a hyperspherical probability distribution)~\cite{wang2020understanding} so that continuous latent states are distributed on the hypersphere.

DreamingV2 defines the critic as categorical PFMs and employs simple parameterization like Eq.~(\ref{eqn:bilinear}).
\texttt{calc\_pmf()} in Fig.~\ref{fig:arch_and_code}(c) describes the pseudo-code for estimating the PMFs' parameters, where the probability matrix is estimated with simple linear transformation $W\boldsymbol{e}_{t}$ and softmax operation.
The logits for contrastive learning can be simply computed like in \texttt{calc\_logit()} in Fig.~\ref{fig:arch_and_code}(c).

\subsection{Auxiliary contrastive objective} \label{sec:aux_obj}
We encountered the issue that just optimizing $J^{\mathrm{Dreaming}}$ did not work (optimization of $J_{k}^{\mathrm{NCE}}$ did not progress at all).
To solve this, we heuristically added the auxiliary objective:
\begin{align}
  J_{\mathrm{aux}}^{\mathrm{NCE}} =
  \mathbb{E}_{p(\latent_{t}|\obs_{t})}
  \left[
    \log \frac{{p(\latent_{t}|\obs_{t})}}{\sum_{\obs'}p(\latent_{t}|\obs_{t})}
  \right] \label{eqn:aux_nce_loss}.
\end{align}
Fig.~\ref{fig:arch_and_code}(b) illustrates the architecture to calculate this objective, which is similar to Momentum Contrast (MoCo) for unsupervised visual representation learning~\cite{he2020momentum}.
This objective contributes to the training the CNN encoder and critic parameter $W$.
We found that this simpler objective than $\nceloss_{k}$ successfully triggers the overall optimization.

\subsection{Momentum encoder}
We used a momentum encoder, which is a general scheme introduced in many representation learning publications~\cite{he2020momentum,grill2020bootstrap,srinivas2020curl,paster2021blast}.
As shown in Fig.~\ref{fig:arch_and_code}(a/b), the stop-gradient operator is inserted after the \texttt{CNN}', and its parameter $\bar{\theta}$ is updated using the exponential moving average of the \texttt{CNN}'s parameter $\theta$; i.e.,~$\bar{\theta} \leftarrow (1 - \eta) \bar{\theta} + \eta \theta$, where $\eta \in (0, 1]$ is a momentum coefficient.
We adopt the momentum encoder since the stop-gradient operation can drastically reduce memory usage.
% \footnote{
Without the operation, the memory complexity of contrastive learning is $O((B \times K)^{2})$ due to the pairwise logit computations, however, it can be reduced to $O(B \times K)$ by the stop-gradient.
% }.

\subsection{Random image augmentation} \label{sec:random_aug}
As in Dreaming, we apply random image augmentation to observations.
A recently introduced random crop method in \cite{yarats2021mastering} is adopted to DreamingV2 since the literature has reported performance improvements on model-free reinforcement learning in pixels.
This augmentation firstly pads each side of a $64\times 64$ observation by 4 pixels (by repeating
boundary pixels), secondly selects a random $64\times 64$ boundary box, and then finally outputs the original image shifted by
4 pixels.

\subsection{Implementation}
We implemented DreamingV2 in TensorFlow based on the official source code of DreamerV2%
\footnote{
  \url{https://github.com/danijar/dreamerv2}
}.
We kept the original policy optimization method, hyperparameters and experimental conditions similar to
the original ones.
DreamingV2 specific hyperparameters $K$ (prediction horizon) and $\eta$ (momentum coefficient)
were set to be $K=3$ and $\eta=0.05$, respectively.
The effect of $K$ is tested in the ablation study in Sec.~\ref{tab:ablation_os}.
We observed no significant effect of $\eta$ as long as the momentum encoder was slowly updated.

\begin{table*}[t]
  \caption{Ablation study: the effect of discrete latent and reconstruction-free world modeling. Boldface and underlines indicate the best results. Underlines mean the second-best.}
  \label{tab:ablation_discrete_recon}
  % \resizebox{0.46\textwidth}{!}{
    \centering
    \begin{tabular}{l|cccc|cc}
      \toprule
      & DreamingV2 (ours) & DreamerV2~\cite{hafner2020mastering} 
      & Dreaming~\cite{okada2021dreaming} & Dreamer~\cite{hafner2019dreamer} 
      & DreamerPro~\cite{deng2021dreamerpro} & TPC~\cite{nguyen2021temporal} \\
      \midrule
      Discrete latent & $\checkmark$ & $\checkmark$ & &  \\
      Reconstruction free & $\checkmark$ & & $\checkmark$ & & $\checkmark$ & $\checkmark$ \\
      \midrule
      \multicolumn{5}{l}{\textbf{(A) 3D Rotot-arm tasks}} \\
      \midrule
      UR5-reach & \best{776}{194} & \third{704}{222} & \second{752}{1178} & \third{701}{223} & \third{668}{252} & \third{555}{200}  \\
      Reach-duplo & \best{199}{43} & \second{149}{62} & \third{145}{61} & \third{5}{11} & \third{87}{76} & \third{5}{16}  \\
      Lift & \best{327}{150} & \third{165}{126} & \second{174}{107} & \third{134}{46} & \third{138}{64} & \third{24}{35} \\
      Door & \best{383}{143} & \third{190}{126} & \second{319}{173} & \third{154}{32} & \third{111}{110} & \third{118}{129} \\
      PegInHole & \best{436}{26} & \second{376}{59} & \third{353}{50} & \third{354}{47} & \third{327}{43} & \third{369}{67} \\
      \midrule
      \midrule
      \multicolumn{5}{l}{(B) 2D Manipulation tasks} \\
      \midrule
      Reacher-hard & \second{598}{447} & \third{175}{340} & \best{743}{346} & \third{247}{392} & - & - \\
      Finger-turn-hard & \third{484}{434} & \second{600}{417} & \best{858}{210} & \third{533}{426} & - & - \\
      % \midrule
      % \multicolumn{5}{l}{(C) 2D Manipulation tasks where object vanishing is NOT critical} \\
      % \midrule
      Reacher-easy & \second{924}{210} & \third{923}{215} & \best{947}{100} & \third{658}{429} & - & - \\
      Finger-turn-easy & \third{434}{469} & \third{498}{469} & \best{842}{286} & \second{665}{430} & - & - \\
      \midrule
      \multicolumn{5}{l}{(C) Difficult pole-swingup tasks} \\
      \midrule
      Acrobot-swingup & \best{470}{129} & \third{309}{131} & \third{359}{111} & \second{382}{147} & - & - \\
      Cartpole-two-poles & \best{308}{55} & \third{248}{103} & \second{273}{53} & \third{256}{65} & - & - \\
      \midrule
      \multicolumn{5}{l}{(D) 3D locomotion robot tasks} \\
      \midrule
      Quadruped-walk & \best{492}{127} & \third{350}{89} & \second{379}{189} & \third{242}{120} & - & - \\
      Quadruped-run & \best{385}{91} & \second{352}{68}  & \third{339}{128} & \third{269}{114} & - & - \\
      \midrule
      \multicolumn{5}{l}{(E) 2D locomotion tasks} \\
      \midrule
      Cheetah-run & \third{768}{24} & \best{811}{75} & \third{542}{132} & \second{776}{120} & - & - \\
      Walker-walk & \third{857}{115} & \best{951}{28} & \third{518}{76} & \second{906}{70} & - & - \\
      \bottomrule
    \end{tabular}
  % }
\end{table*}

\section{Experiments} \label{sec:experiments}
\subsection{Comparison to state-of-the-art methods}
The main objective of this experiment is to demonstrate that \proposedmethod{} has advantages over the baseline method
DreamerV2~\cite{hafner2020mastering} on five robot-arm tasks shown in the top part of Fig.~\ref{fig:main_result}.
In contrast to the commonly used 2D robot tasks in the DeepMind Control Suite (e.g., Cheetah, Walker)~\cite{deepmindcontrolsuite2018}, these tasks must be solved by understanding complex vision observations from 3D spaces.
In addition, iGibsonRenderer~\cite{li2021igibson} is introduced for high-quality visualization of RoboSuite tasks (i.e.,~{Lift}, {Door}, {PegInHole}).
% This makes autoencoding-based world models to reconstruct such the complex observations.
Although these tasks are in virtual environments, we assume that they well simulate real-world scenarios.
This experiment includes the recent world model based reinforcement learning without reconstruction,
DreamerPro~\cite{deng2021dreamerpro} and TPC~\cite{nguyen2021temporal}%
\footnote{
Code was taken from \url{https://github.com/fdeng18/dreamer-pro}
}.
The bottom part of Fig.~\ref{fig:main_result} shows the training curves of the experiments.
In these results, DreamingV2 consistently outperformed other baselines.
In addition, on the tasks of Reach-duplo, {Lift}, {Door}, and {PegInHole}, DreamingV2 significantly showed better performances. Moreover, only DreamingV2 could solve these tasks within 500K($=500\times 10^{3}$) steps, which is about seven hours of interactions with the environments%
\footnote{
  In the Robosuite tasks, control frequency was 20~Hz.
  Hence, 500K steps are approximately equivalent to seven hours of robot operation.
}.
\begin{figure}
  \centering
  \includegraphics[width=0.45\textwidth]{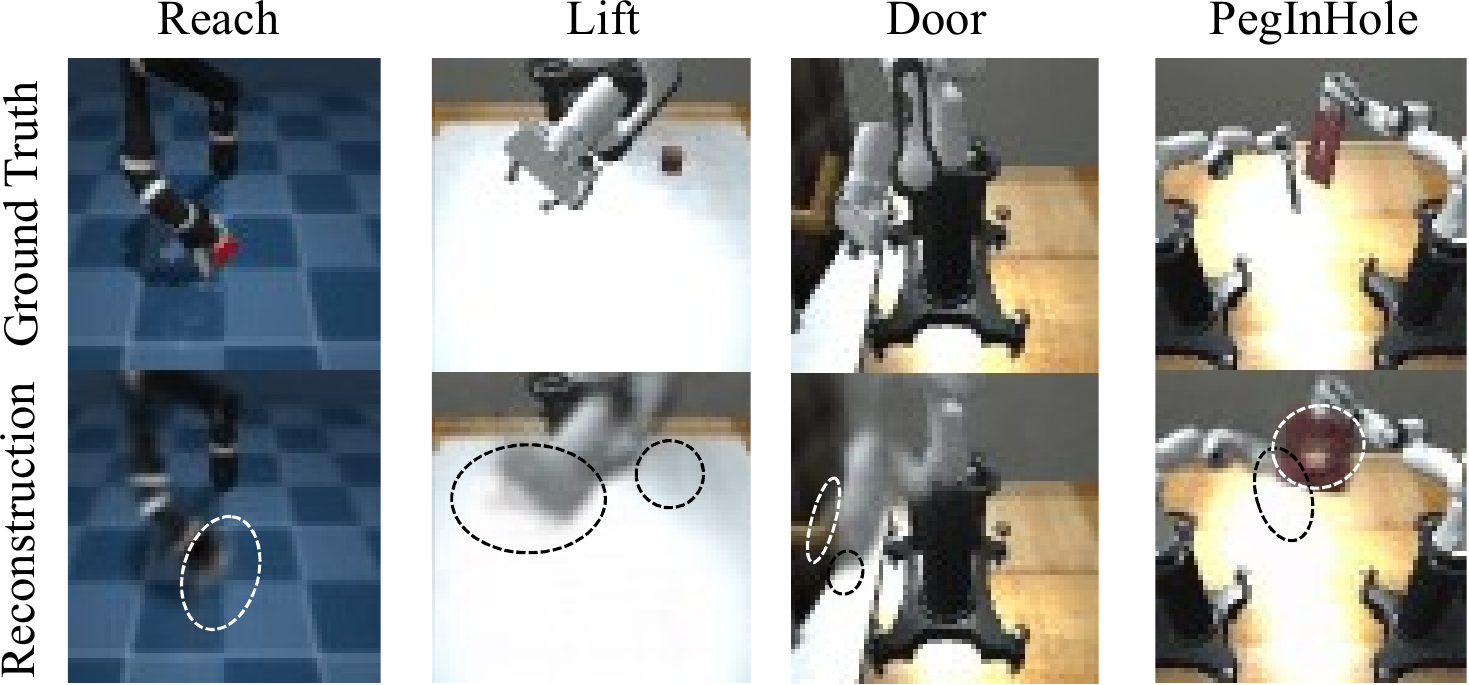}
  \caption{
    Reconstructed images by DreamerV2's world models.
    The dashed-line ovals highlight the \textit{object vanishing}.
  }
  \label{fig:object_vanishing}
\end{figure}

\subsection{Ablation study: continuous latent vs. discrete latent, Reconstruction-based vs. -free learning}
% \subsubsection{}
In contrast to the previous literature, DreamingV2 suppports both discrete latent and reconstruction-free learning.
This ablation study reveals how these factors contributed to the above results.
For this purpose, we compared DreamingV2 with DreamerV2, Dreamer, and Dreaming.
Dreamer and Dreaming are implemented based on the original DreamerV2 and our DreamingV2 code.
Hence, several components introduced in DreamerV2 and DreamingV2 (i.e., KL-balancing, policy entropy regularization, momentum encoder, and auxiliary contrastive loss), are inherited by Dreamer and Dreaming.
These changes from the original implementation of Dreamer and Dreaming resulted in several performance changes from \cite{okada2021dreaming}, but we left them untouched to evaluate the effect of discrete latent and reconstruction-free learning.
In addition to the previous robot arms tasks, another variety of 10 DMC tasks are evaluated, which are categorized into four classes namely; 2D manipulation, pole-swingup, 3D locomotion, and 2D locomotion.

Table~\ref{tab:ablation_discrete_recon} summarizes the training results benchmarked at 500K steps.
The results show the mean and standard deviation averaged eight random seed experiments.
\textbf{(A) On the robot arm tasks}, DreamingV2 consistently outperformed other Dreamer-variants.
We observed reconstruction-based method suffered from reconstructing complex vision observations as shown in Fig.~\ref{fig:object_vanishing}.
Dreaming also performed well and ranked second on UR5-reach, Lift, and Door tasks.
This also shows the effectiveness of reconstruction-free learning on complex observation tasks.
However, Dreaming performed significantly worse than DreamingV2 on tasks other than UR5-reach.
Because these tasks require contact-rich manipulations, the resulting dynamics are highly discontinuous and difficult to represent with continuous latent variables.
% In addition, \proposedmethod{} achieves outperforming performance than the leading MFRL methods.
% We also compare \proposedmethod{} and Dreamer on another variety of 10 DMC-tasks,
% which are selected mainly because of difficulty from reward sparcity and/or high-dimensionality of state-action space.
% Only CURL is evaluated for the additional 10 tasks.
\textbf{(B) On the 2D manipulation tasks}, Dreaming demonstrated the best performances.
Since these simple environments can be almost described by angular information (joint and rotater angles), we assume that Dreaming with continuous latent on the hypersphere (see Sec.~\ref{sec:critic}) took advantages to represent these polar coordinates worlds.
In contrast, DreamingV2 struggled represent these environments with discrete latent, showing worse performance especially in Finger tasks.
% While, discrete latent is less effective to describe such the environment, hence DreamingV2 performed worse than Dreaming.
% there are no significant difference between \proposedmethod{} and Dreamer because the key objects are large enough.
\textbf{(C) On the pole swingup tasks}, DreamingV2 achieved the best performance.
Despite their low degrees of freedom, these tasks take a wide range of states and provide a diverse of observations, making reconstruction challenging.
The higher performance of DreamingV2 over Dreaming indicates the effectiveness of discrete representation to such chaotic systems~\cite{ueda2008dynamic}.
\textbf{(D) On 3D locomotion tasks}, DreamingV2 was ranked best.
As similar with the robot arm tasks, 3D observations are complex to reconstruct, limiting the performance of Dreamer and DreamerV2.
Since these tasks need contact-rich controls, DreamingV2 performed better than Dreaming.
\textbf{(E) On 2D locomotion tasks}, DreamerV2 outperformed other methods.
As reported in \cite{okada2021dreaming}, contrastive learning fails to capture the velocity information from \textit{robot-centric} observations. However, by capturing discontinuous dynamics with discrete representation, DreamingV2 significantly improved the performance than Dreaming.
% We also compare the performance with the previous versions of the proposed method and the main baseline, Dreaming~\cite{okada2021dreaming} and Dreamer~\cite{hafner2019dreamer}.
%

\subsection{Ablation study: hyper parameters}
\subsubsection{The effect of multi-step prediction horizon $K$} \label{sec:ablation_k}
Table~\ref{tab:ablation_os} analyzes the effect of prediction horizon $K$ on the robot arm tasks.
In almost cases, the longer prediction showed better performance, but it saturated at values of $K=2$ and $K=3$.
\begin{table}[tb]
  \caption{
  Ablation study: The effect of the overshooting distance $K$.
  }
  \label{tab:ablation_os}
  \centering
  \resizebox{0.45\textwidth}{!}{
%   {\small
  \begin{tabular}{l|cccc}
    \toprule
    & $K=0$ & $K=1$ & $K=2$ & $K=3$ \\
    \midrule
    UR5-reach & \best{783}{190} & \third{686}{229} & \third{706}{232} & \second{776}{194} \\
    Reach-duplo & \third{157}{61} & \third{169}{52} & \second{187}{54} & \best{199}{43} \\ %& 631 $\pm$ \scriptsize{397} \\
    Lift & \third{197}{83} & \third{273}{162} & \best{353}{131} & \second{327}{150}
    \\ %& 11 $\pm$ \scriptsize{45} \\
    Door & \third{323}{172} & \third{369}{154} & \best{424}{81} & \second{383}{143}
    \\
    PegInHole & \third{398}{53} & \third{421}{46} & \second{434}{52} & \best{436}{26}
    \\
    \bottomrule
  \end{tabular}
  }
\end{table}

% \begin{figure}
%   \centering
%   \includegraphics[width=0.35\textwidth]{fig/ablation_k-crop.pdf}
%   \caption{Ablation study: the effect of the overshooting distance $K$.}
%   \label{fig:ablation_k}
% \end{figure}

% This experiment is conducted to analyze how the major components of the proposed representation learning, introduced in Sec.~\ref{sec:instantiation}, contribute to the overall performance.
\subsubsection{Modeling of the auxiliary dynamics $\tilde{p}$} \label{sec:ablation_aux_dyn}
Table~\ref{tab:ablation_linearity} probes the effect of the auxiliary dynamics modeling (linear as in Dreaming~\cite{okada2021dreaming} vs. RSSM).
RSSM modling performed better on 3 of 5 tasks, which are different results from \cite{okada2021dreaming} because we employed discrete representation.
Linear modeling demonstrated better on 2 of 5 tasks, however, the differences seemed not significant.
% \begin{figure}
%   \centering
%   \includegraphics[width=0.35\textwidth]{fig/ablation_aux_dyn-crop.pdf}
%   \caption{Ablation study: the effect of the auxiliary dynamics.}
%   \label{fig:ablation_aux_dyn}
% \end{figure}
\begin{table}[t]
  \caption{
    Ablation study: The effects of the auxiliary dynamics.
  }
  \label{tab:ablation_linearity}
  \centering
  \begin{tabular}{l|cc}
    \toprule
                          & \multicolumn{1}{c}{$\tilde{p}$: linear as in \cite{okada2021dreaming}}
                          & \multicolumn{1}{c}{$\tilde{p}$: RSSM}
                          % & \multirow{2}{*}{-}
                          \\
    \midrule
    UR5-reach & \second{788}{154} & \best{834}{194} \\
    Reach-duplo & \second{210}{66} & \best{219}{43} \\
    Lift & \second{319}{151} & \best{396}{150} \\
    Door & \best{462}{141} & \second{450}{143}   \\
    PegInHole & \best{437}{50} & \second{430}{26}  \\
    \bottomrule
  \end{tabular}
\end{table}

\subsection{Analysis of latent states} \label{sec:analysis_latent}
\begin{figure}
  \centering
  \includegraphics[width=0.45\textwidth]{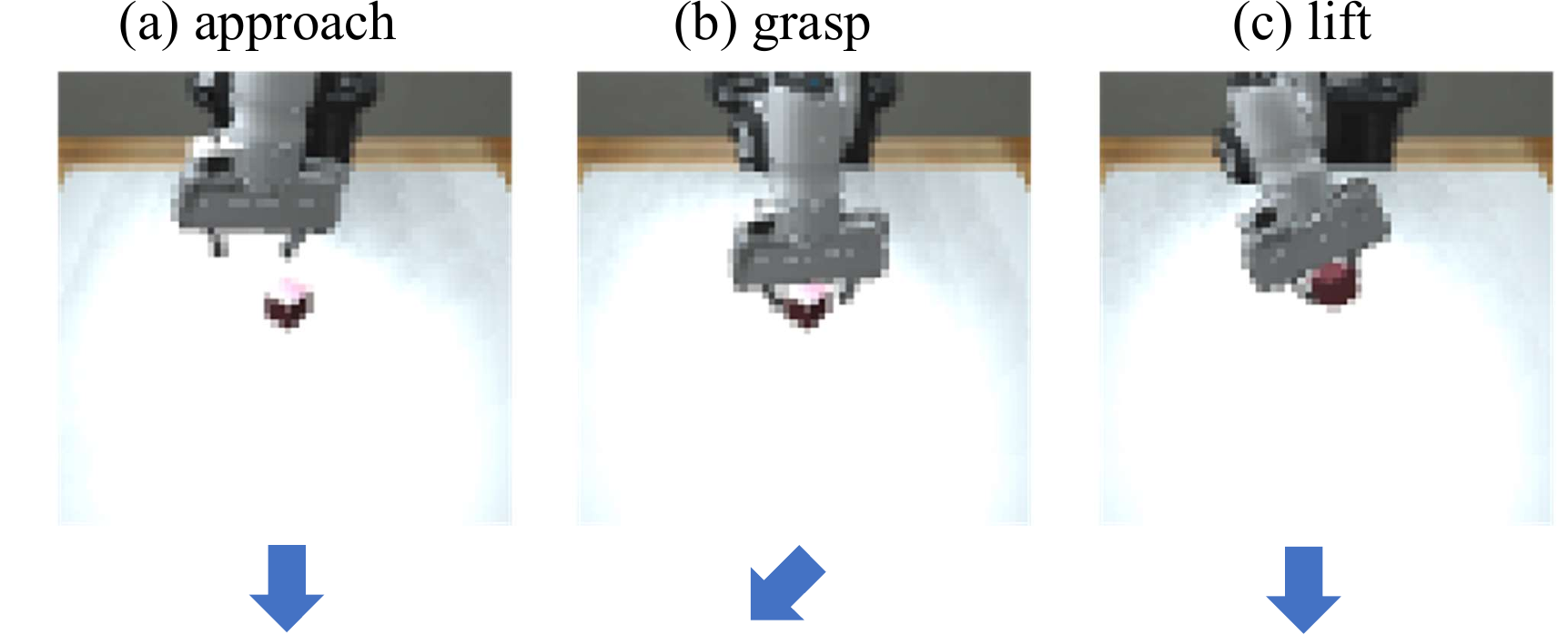} \\
  % \vspace{5mm}
  \includegraphics[width=0.4\textwidth]{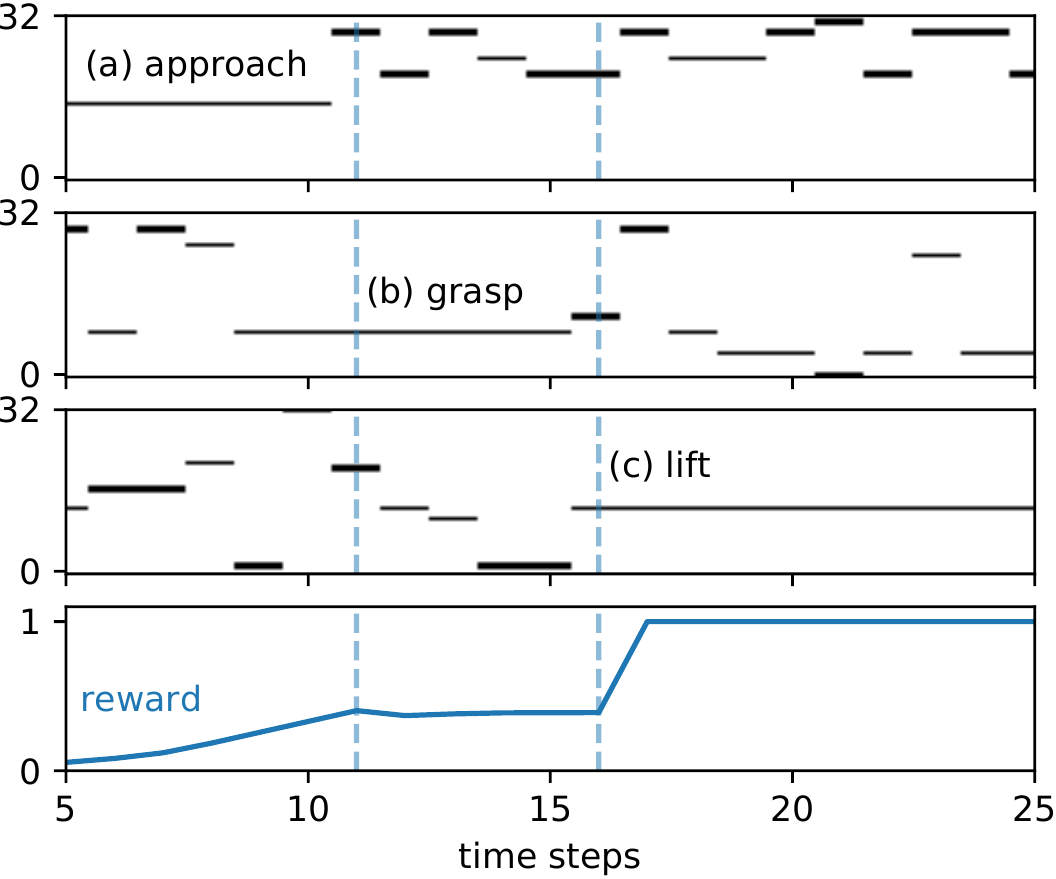}
  \caption{
    Latent and reward trajectories of DreamingV2 during a test trial on Lift task.
    Only 3 of 32 categorical variables, which were supposed to represent subtask phases,
    were picked up.
  } \label{fig:analysis_latent}
\end{figure}

Fig.~\ref{fig:analysis_latent} visualizes latent trajectories on Lift task generated by DreamingV2,
in which we can observe three categorical latent variables that represent tasks phases, i.e., (a) approach, (b) grasp, and (c) lift.

%
% \begin{figure}
%   \centering
%   \begin{minipage}{0.23\textwidth}
%     \centering
%     \includegraphics[width=\textwidth]{fig/hist_lift_baseline.pdf}
%     {\footnotesize (a) DreamerV2: $\mathcal{H}=2.$}
%   \end{minipage}
%   \begin{minipage}{0.23\textwidth}
%     \centering
%     \includegraphics[width=\textwidth]{fig/hist_lift_ours.pdf}
%         {\footnotesize (b) DreamingV2: $\mathcal{H}=2.$}
%   \end{minipage}
%   \caption{
%     Empirical latent distributions and entropies $\mathcal{H}$ from 100 test trials of .
%   }
% \end{figure}

\section{Related Work} \label{sec:related_work}
% \textbf{Self-supervised methods for image representation learning: }
% Contrastive learning is one of the most accepted strategies for self-supervised learning~\cite{he2020momentum,chen2020simple,chen2021exploring}.
% especially in image processing domains
%  natural language processing~\cite{giorgi2020declutr}, and audio processing~\cite{saeed2021contrastive}.
% \textbf{Model-free reinforcement learning with self-supervised learning:}
% CURL~\cite{srinivas2020curl} is an application of the contrastive learning for reinforcement learning, but CURL is a model-free method based on Soft Actor Critic (SAC)~\cite{haarnoja2018soft} and the contrastive obejctive is utilized to train only encoders not world models.
% Deep bisimulation for control (DBC)~\cite{zhang2020learning} and
% discriminative particle filter reinforcement learning (DPFRL)~\cite{ma2020discriminative} are other types of cutting edge MFRL methods, which utilize different concepts of representation learning without reconstruction.

% % \textbf{World models with reconstruction:}
% Stochastic Latent Actor Critic (SLAC)~\cite{lee2019stochastic} learns a policy in a model-free fashion but optimizes general world model objective similar to Eq.~\ref{eqn:rssm} as an auxiliary objective to facilitate policy evolution in pixel based objective tasks.
Recently, world models with self-supervised learning have received a lot of attention, and several methods for reconstruction-free model-based learning have been developed.
CFM (Contrastive Forward Model)~\cite{yan2020learning} utilizes a similar reconstruction-free objective like Dreaming(V2), however, it is dedicated to continuous representation.
CVRL~\cite{ma2020contrastive}, TPC~\cite{nguyen2021temporal}, and Multi-State Space Model (MSSM)~\cite{chen2021multi} also introduce contrastive learning for decoder-free world modeling.
The major differences between them and DreamingV2 are continuous latent representation and the lack of multi-step prediction for contrastive learning (i.e., $K=0$).
The specific differences that characterize these methods are; CVRL introduces an auxiliary objective for policy optimization and world-model predictive control at test time, TPC replaces the critic $p(\latent_{t}|\obs_{t})$ with $q(\latent_{t}|\latenth_{t},\obs_{t})$, and MSSM computes logits by comparing a pair of latent variables originating from different modals (e.g., different camera views).
DreamerPro~\cite{deng2021dreamerpro} introduces the clustering-based or prototype-based method~\cite{caron2020unsupervised}, in which a set of clusters represented by finite quantized vectors are trained, thereby eliminating pair-wise comparisons of contrastive learning.
Bootstrapped LAtents for Simulating Trajectories (BLAST)~\cite{paster2021blast} uses heuristics from the recent self-supervised method for vision, Bootstrap Your Own Latent (BYOL)~\cite{grill2020bootstrap}, to achieve representation learning without reconstruction and negative samples.

\section{Conclusion} \label{sec:conclusion}
% Newtonian VAE~\cite{jaques2021newtonianvae}, locally linear~\cite{watter2015embed}
% Barlow Twins~\cite{zbontar2021barlow},
In the present paper, we proposed DreamingV2, a synergistic extention of DreamerV2 and Dreaming.
DreamingV2 is equipped with discrete latent representation and reconstruction-free objective, which differentiate our method from other existing literature of world models.
To the best of our knowledge, applying contrastive learning to discrete latent is the world's first challenge, and we successfully demonstrated that our method could facilitate policy evolution to solve the 3D, visually complex, and contact-rich robot-arm tasks from pixel observations.
This result strongly suggests the applicability of industrial use of this method.
We believe that this method is also applicable for other complex observation tasks, such as point cloud~\cite{liu20173dcnn}, and multimodal inputs~\cite{chen2021multi}.

Although we demonstrated the effectiveness of discrete representation in the robot-arm tasks,
DreamingV2 performed worse scores than other methods based on continuous representation on several tasks like Finger-turn.
The task dependency of representation needs to be considered in detail in the future study.
Hybridization of continuous and discrete representations appears to be attractive for future direction.
In addition, another challenge of the proposed method is the increase in latent dimensionality due to the discrete representation
(Dreaming and DreamingV2 respectively require $32$- and $1024(=32 \times 32)$-dimensional vectors for $\latent$).
% This increase in dimensionality is especially critical in contrast learning because the pairwise logit computation require a  complexity proportional to the square of the sample numberin mini-batch.
Our method runs on a single NVIDIA V100 GPU, but the high dimensionality limits the computation speed (approximately 1.7x slower than DreamerV2).
A solution to this is to use negative sample free self-supervised learning such as BYOL~\cite{grill2020bootstrap} and  Barlow Twins~\cite{zbontar2021barlow}.

\section*{ACKNOWLEDGMENT}
Most of the experiments were conducted in ABCI (\textit{AI Bridging Cloud Infrastructure}), built by the National Institute of Advanced Industrial Science and Technology, Japan.

% \clearpage
\bibliography{iros2022}

\begin{thebibliography}{10}

\bibitem{ha2018world}
D.~Ha and J.~Schmidhuber, ``World models,'' {\em arXiv:1803.10122}, 2018.

\bibitem{watter2015embed}
M.~Watter, J.~Springenberg, J.~Boedecker, and M.~Riedmiller, ``Embed to
  control: A locally linear latent dynamics model for control from raw
  images,'' {\em NeurIPS}, 2015.

\bibitem{hafner2018learning}
D.~Hafner, T.~Lillicrap, I.~Fischer, R.~Villegas, D.~Ha, H.~Lee, and
  J.~Davidson, ``Learning latent dynamics for planning from pixels,'' in {\em
  ICML}, 2019.

\bibitem{okada2020planet}
M.~Okada, N.~Kosaka, and T.~Taniguchi, ``{PlaNet} of the {Bayesians}:
  Reconsidering and improving deep planning network by incorporating {Bayesian}
  inference,'' in {\em IROS}, 2020.

\bibitem{hafner2019dreamer}
D.~Hafner, T.~Lillicrap, J.~Ba, and M.~Norouzi, ``Dream to control: Learning
  behaviors by latent imagination,'' {\em ICLR}, 2020.

\bibitem{hafner2020mastering}
D.~Hafner, T.~Lillicrap, M.~Norouzi, and J.~Ba, ``Mastering atari with discrete
  world models,'' in {\em ICLR}, 2021.

\bibitem{okada2021dreaming}
M.~Okada and T.~Taniguchi, ``Dreaming: Model-based reinforcement learning by
  latent imagination without reconstruction,'' in {\em ICRA}, 2021.

\bibitem{byravan2020imagined}
A.~Byravan, J.~T. Springenberg, A.~Abdolmaleki, R.~Hafner, {\em et~al.},
  ``Imagined value gradients: Model-based policy optimization with tranferable
  latent dynamics models,'' in {\em CoRL}, 2020.

\bibitem{sekar2020planning}
R.~Sekar, O.~Rybkin, K.~Daniilidis, P.~Abbeel, D.~Hafner, and D.~Pathak,
  ``Planning to explore via self-supervised world models,'' {\em
  arXiv:2005.05960}, 2020.

\bibitem{yu2020mopo}
T.~Yu, G.~Thomas, L.~Yu, S.~Ermon, J.~Y. Zou, S.~Levine, C.~Finn, and T.~Ma,
  ``{MOPO}: Model-based offline policy optimization,'' {\em NeurIPS}, 2020.

\bibitem{sakai2022explainable}
T.~Sakai and T.~Nagai, ``Explainable autonomous robots: a survey and
  perspective,'' {\em Advanced Robotics}, pp.~1--20, 2022.

\bibitem{deepmindcontrolsuite2018}
Y.~Tassa, Y.~Doron, A.~Muldal, T.~Erez, Y.~Li, {\em et~al.}, ``Deep{Mind}
  control suite,'' {\em arXiv:1801.00690}, 2018.

\bibitem{bellemare2013arcade}
M.~G. Bellemare, Y.~Naddaf, J.~Veness, and M.~Bowling, ``The arcade learning
  environment: An evaluation platform for general agents,'' {\em Journal of
  Artificial Intelligence Research}, vol.~47, pp.~253--279, 2013.

\bibitem{okada2019variational}
M.~Okada and T.~Taniguchi, ``Variational inference {MPC} for bayesian
  model-based reinforcement learning,'' in {\em CoRL}, 2019.

\bibitem{ma2020contrastive}
X.~Ma, S.~Chen, D.~Hsu, and W.~S. Lee, ``Contrastive variational model-based
  reinforcement learning for complex observations,'' {\em CoRL}, 2020.

\bibitem{nguyen2021temporal}
T.~D. Nguyen, R.~Shu, T.~Pham, H.~Bui, and S.~Ermon, ``Temporal predictive
  coding for model-based planning in latent space,'' in {\em ICML}, 2021.

\bibitem{deng2021dreamerpro}
F.~Deng, I.~Jang, and S.~Ahn, ``{DreamerPro}: Reconstruction-free model-based
  reinforcement learning with prototypical representations,'' {\em
  arXiv:2110.14565}, 2021.

\bibitem{oord2018representation}
A.~v.~d. Oord, Y.~Li, and O.~Vinyals, ``Representation learning with
  contrastive predictive coding,'' {\em arXiv:1807.03748}, 2018.

\bibitem{chen2020simple}
T.~Chen, S.~Kornblith, M.~Norouzi, and G.~Hinton, ``A simple framework for
  contrastive learning of visual representations,'' in {\em ICLR}, 2020.

\bibitem{han2019variational}
D.~Han, K.~Doya, and J.~Tani, ``Variational recurrent models for solving
  partially observable control tasks,'' in {\em ICLR}, 2020.

\bibitem{cho2014learning}
K.~Cho, B.~Van~Merri{\"e}nboer, C.~Gulcehre, D.~Bahdanau, {\em et~al.},
  ``Learning phrase representations using {RNN} encoder-decoder for statistical
  machine translation,'' {\em arXiv:1406.1078}, 2014.

\bibitem{zhu2020robosuite}
Y.~Zhu, J.~Wong, A.~Mandlekar, and R.~Mart{\'\i}n-Mart{\'\i}n, ``robosuite: A
  modular simulation framework and benchmark for robot learning,'' {\em
  arXiv:2009.12293}, 2020.

\bibitem{li2021igibson}
C.~Li, F.~Xia, R.~Mart\'in-Mart\'in, M.~Lingelbach, S.~Srivastava, {\em
  et~al.}, ``{iGibson} 2.0: Object-centric simulation for robot learning of
  everyday household tasks,'' {\em arXiv:2108.03272}, 2021.

\bibitem{bengio2013estimating}
Y.~Bengio, N.~L{\'e}onard, and A.~Courville, ``Estimating or propagating
  gradients through stochastic neurons for conditional computation,'' {\em
  arXiv:1308.3432}, 2013.

\bibitem{wang2020understanding}
T.~Wang and P.~Isola, ``Understanding contrastive representation learning
  through alignment and uniformity on the hypersphere,'' in {\em ICLR}, 2020.

\bibitem{he2020momentum}
K.~He, H.~Fan, Y.~Wu, S.~Xie, and R.~Girshick, ``Momentum contrast for
  unsupervised visual representation learning,'' in {\em CVPR}, 2020.

\bibitem{grill2020bootstrap}
J.-B. Grill, F.~Strub, F.~Altch{\'e}, C.~Tallec, {\em et~al.}, ``Bootstrap your
  own latent-a new approach to self-supervised learning,'' {\em NeurIPS}, 2020.

\bibitem{srinivas2020curl}
A.~Srinivas, M.~Laskin, and P.~Abbeel, ``{CURL}: Contrastive unsupervised
  representations for reinforcement learning,'' in {\em ICML}, 2020.

\bibitem{paster2021blast}
K.~Paster, L.~E. McKinney, S.~A. McIlraith, and J.~Ba, ``{BLAST}: Latent
  dynamics models from bootstrapping,'' in {\em Deep RL Workshop NeurIPS 2021},
  2021.

\bibitem{yarats2021mastering}
D.~Yarats, R.~Fergus, A.~Lazaric, and L.~Pinto, ``Mastering visual continuous
  control: Improved data-augmented reinforcement learning,'' in {\em ICLR},
  2022.

\bibitem{ueda2008dynamic}
R.~Ueda and T.~Arai, ``Dynamic programming for global control of the acrobot
  and its chaotic aspect,'' in {\em ICRA}, 2008.

\bibitem{yan2020learning}
W.~Yan, A.~Vangipuram, P.~Abbeel, and L.~Pinto, ``Learning predictive
  representations for deformable objects using contrastive estimation,'' {\em
  arXiv:2003.05436}, 2020.

\bibitem{chen2021multi}
K.~Chen, Y.~Lee, and H.~Soh, ``Multi-modal mutual information ({MUMMI})
  training for robust self-supervised deep reinforcement learning,'' in {\em
  ICRA}, 2021.

\bibitem{caron2020unsupervised}
M.~Caron, I.~Misra, J.~Mairal, P.~Goyal, P.~Bojanowski, and A.~Joulin,
  ``Unsupervised learning of visual features by contrasting cluster
  assignments,'' {\em NeurIPS}, 2020.

\bibitem{liu20173dcnn}
F.~Liu, S.~Li, L.~Zhang, C.~Zhou, R.~Ye, Y.~Wang, and J.~Lu, ``{3DCNN-DQN-RNN}:
  A deep reinforcement learning framework for semantic parsing of large-scale
  3d point clouds,'' in {\em CVPR}, 2017.

\bibitem{zbontar2021barlow}
J.~Zbontar, L.~Jing, I.~Misra, Y.~LeCun, and S.~Deny, ``{Barlow Twins}:
  self-supervised learning via redundancy reduction,'' in {\em ICML}, 2021.

\end{thebibliography}
\bibliographystyle{ieeetr}

\end{document}